\documentclass[lettersize,journal]{IEEEtran}
\usepackage{amsmath,amsfonts}
\usepackage{url}
\usepackage{adjustbox}
\usepackage{algorithm}
\usepackage{algpseudocode}
\usepackage{amsmath}
\usepackage{pifont}
\usepackage{booktabs}
\usepackage{tabularx}
\usepackage{multirow}
\usepackage{tabularx}

\usepackage{graphicx}
\usepackage{subcaption} 
\usepackage{float}
\usepackage{threeparttable}
\usepackage{array}
\usepackage{multirow}
\usepackage{array}

\usepackage{xltabular}
\usepackage{notoccite}
\usepackage[noadjust]{cite}
\usepackage{stfloats}
\usepackage{url}
\usepackage{cuted}
\usepackage{caption}

\begin{document}

\title{MRPoS: Mixed Reality-Based Robot Navigation Interface Using Spatial Pointing and Speech with Large Language Model}
\author{
Eduardo Iglesius$^{\dag 1}$, Masato Kobayashi$^{\dag 1,2*}$, Yuki Uranishi$^{1}$
\thanks{
$^{\dag}$ Equal Contribution, 
$^{1}$ The University of Osaka, 
$^{2}$ Kobe University,\\
$^*$ corresponding author: kobayashi.masato.cmc@osaka-u.ac.jp
}
}

\maketitle
\begin{abstract}
Recent advancements have made robot navigation more intuitive by transitioning from traditional 2D displays to spatially aware Mixed Reality (MR) systems. However, current MR interfaces often rely on manual "air tap" gestures for goal placement, which can be repetitive and physically demanding, especially for beginners. This paper proposes the Mixed Reality-Based Robot Navigation Interface using Spatial Pointing and Speech (MRPoS). This novel framework replaces complex hand gestures with a natural, multimodal interface combining spatial pointing with Large Language Model (LLM)-based speech interaction. By leveraging both information, the system translates verbal intent into navigation goals visualized by MR technology. Comprehensive experiments comparing MRPoS against conventional gesture-based systems demonstrate that our approach significantly reduces task completion time and workload, providing a more accessible and efficient interface.
For additional material, please check: \url{https://mertcookimg.github.io/mrpos}
\end{abstract}

\section{Introduction}

Autonomous mobile robots (AMRs) have emerged as a transformative technology within domestic, healthcare, and industrial fields. Notably, with the advent of the Robot Operating System (ROS) \cite{ros}, robotic navigation has been highly abstracted, requiring users only to input a target pose, and the system will autonomously navigates to the target.

Current tools for generating and visualizing goal poses typically rely on these modalities: 2D computer displays, natural language (speech or text), or MR gestures. However, each approach possesses inherent limitations. Conventional 2D interfaces reduce operational efficiency by forcing users to constantly shift their attention between a screen and the real environment. Language-centric interfaces, despite leveraging LLM for high-level reasoning, struggle with spatial grounding \cite{LLMLimitation}, as describing precise coordinate and orientation often introduces ambiguity. Recent works attempt to bridge this perception-action gap using Vision-Language-Action (VLA) models \cite{omnivla}. Conversely, while MR offers spatial precision necessary to define target poses accurately, relying exclusively on mid-air hand gestures to specify target poses for too long induces a high overall workload \cite{MRLimitation}. 

\begin{figure}[t]
    \centering
    \includegraphics[width=.97\columnwidth]{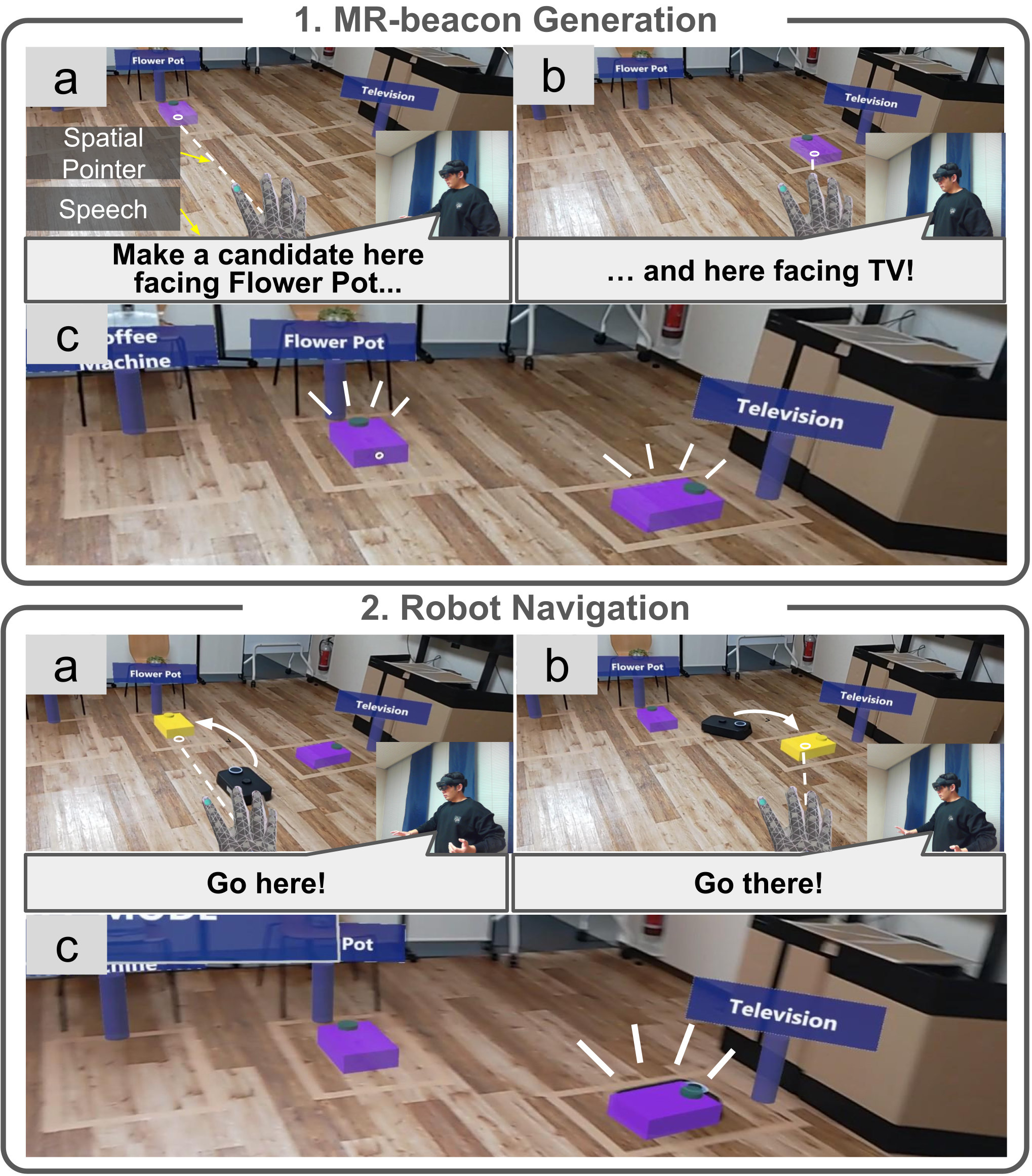}
    
    \caption{\textbf{MRPoS Workflow}. The workflow is divided into 2 phases, which are MR-beacon Generation and Robot Navigation. (1-a, 1-b) shows the MR-beacon generation by using spatial pointer and voice. (1-c) shows the final state after the generation. (2-a, 2-b) shows the robot navigation towards the created MR-beacon. (2-c) shows the final state.}
    \label{fig:concept}
\end{figure}

To address these limitations, this paper introduces MRPoS, a novel framework that optimizes goal pose generation by fusing speech with MR. We propose that these modalities are complementary: MR mitigates the spatial blindness of LLMs, while the semantic power of natural language reduces the workload inherent in MR manipulation. This hybrid approach allows user to simultaneously define a target location via spatial pointing while specifying the desired orientation through intuitive voice commands (Fig.~\ref{fig:concept}). MRPoS features four core functionalities: Add to place single or multiple MR-beacons, Edit to adjust the MR-beacon's location and rotation, Go to dispatch the robot to the MR-beacon, and Delete to remove an MR-beacon from the environment.

\begin{figure*}[t]
    \centering
    \includegraphics[width=.92\textwidth]{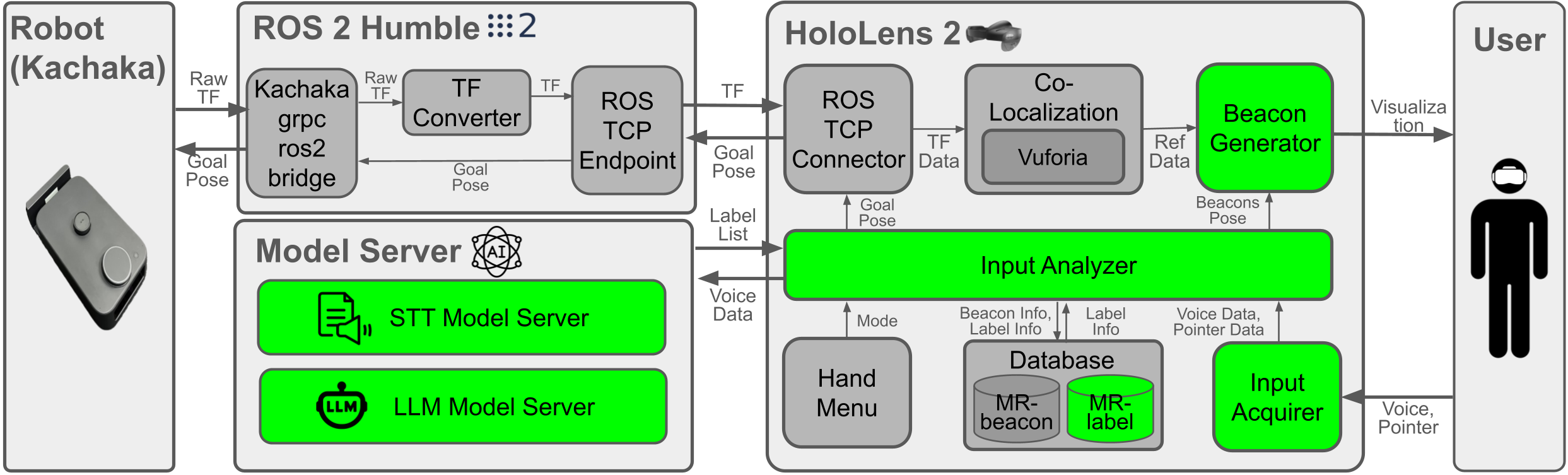}
    \caption{\textbf{System Design Diagram.} Our system consists of four main modules: the HoloLens~2, serving as the user interface to acquire, analyze, and visualize MR-beacons; ROS~2, acting as a bridge between the HoloLens~2 and the robot; a Model Server to process user voice data; and the robot itself. Components highlighted in green represent our novel contributions.}
    \label{fig:SysDesign}
\end{figure*}

Our contributions are summarized as follows:
\begin{itemize}
    \item This paper introduces MRPoS, a multimodal system that streamlines multiple robotic goal poses generation by combining natural language for intuitive interaction with MR spatial pointing for precise spatial grounding.
    \item This paper proposes an architecture that independently interprets voice and pointer inputs before integrating them to calculate the final target pose for multiple goals.
    \item We conducted experiments to compare MRPoS with the baseline gesture-based MR interface, showing the advantages in workload and operational efficiency.
\end{itemize}


\section{Related Work}

\subsection{MR Robot Navigation Interface}

Integrating MR into robot navigation significantly increases operational efficiency by aligning the robot's internal map with the user's physical environment. Applications vary widely, from visualizing motion intent, safety distance \cite{ihandai, intent} to defining restricted zones \cite{mrhad} and manual operation \cite{eye_manual}.

For autonomous navigation, MRNaB~\cite {mrnab}, which is the improvement of  ARviz~\cite {arviz}, enhances traditional 2D SLAM-based occupancy grids by projecting them into the HoloLens~2 environment. This system enables visualization of the goal pose by using a persistent MR-beacon similar to the real object, which projects the navigation goals via ``air tap'' hand gestures. Other studies have also explored the use of discrete waypoints as a primary modality for setting navigation objectives \cite{holo-spok, map3, Drone, sar}. Alternative interfaces include the use of a drag-and-drop interface \cite{holospot, spot-on, DnD} and also eye movement \cite{journal:gaze}.

Despite the advancements, these developments also face several challenges : 
\begin{itemize}
\item Serialized Waypoint Generation: Traditionally, defining a goal pose's position and orientation is a sequential process. This introduces operational latency when generating multiple navigation candidates simultaneously.
\item Elevated Workload: Relying on specialized mid-air hand gestures imposes a steep learning curve which increases ergonomic strain and cognitive demand.
\end{itemize}

This research addresses these limitations by integrating a speech interface with LLMs to reduce workload through natural language interaction. Furthermore, we parallelize these generation process by utilizing natural language for orientation (rotation) and MR spatial pointers for position.

\subsection{Language-Centric Robot Navigation}
LLMs have enabled more natural language interaction for navigation-related tasks, ranging from dialogue-based reasoning and planning \cite{LLMBidirectionalIROS, LLMBidirectional, LLMCollaboration, LLMGuide, LLMambiguity} to language-driven navigation benchmarks and policies \cite{cow, lelan, cast, smartway}.
However, language-only interfaces often face two practical issues in navigation:
(1) limited spatial grounding, where precise goal coordinates and orientations are difficult to specify unambiguously, and
(2) reduced intent transparency, due to the user's inability to verify the target goal pose in the physical environment prior to execution.
Multimodal approaches (e.g., combining language with sketches or visual inputs) partially address these issues \cite{omnivla, LLMSketching}, but their interaction and visualization differ from MR-based goal specification.

In this paper, we treat language primarily as an \emph{auxiliary modality} to reduce interaction burden in the MR interface.
MRPoS combines MR pointing (for spatial grounding) with LLM-processed speech (for expressing orientation) while keeping the goal pose visible and verifiable in situ.
Thus, our evaluation focuses strictly on an MR-to-MR comparison.

\begin{table}[t]
\centering
\scriptsize
\begin{threeparttable}
\caption{Qualitative design-space summary of robotic navigation interface paradigms.}
\label{tab:interface_comparison}

\newcolumntype{Y}{>{\centering\arraybackslash}X}

\begin{tabularx}{\linewidth}{lYYY}
\toprule
\textbf{} & \textbf{Gesture-Based MR} & \textbf{Language-Centric} & \textbf{MRPoS (Hybrid)} \\
\midrule
Natural Language Interaction & \ding{55}  & \checkmark & \checkmark \\
Visual Intent Transparency   & \checkmark & \ding{55}  & \checkmark \\
Precise Spatial Grounding    & \checkmark & \ding{55}  & \checkmark \\
Low Workload                 & \ding{55}  & \checkmark & \checkmark \\
\bottomrule
\end{tabularx}

\begin{tablenotes}[flushleft]
\footnotesize
\item Gesture-based MR refers to \cite{mrhad, arviz, mrnab,  holo-spok, map3, holospot}.
Language-centric refers to \cite{LLMBidirectionalIROS, LLMBidirectional, LLMCollaboration,cow, lelan, llmwaypoint}.
\end{tablenotes}
\end{threeparttable}
\end{table}

\section{System Design}

The system design, illustrated in Fig. \ref{fig:SysDesign}, integrates the Kachaka robot and the HoloLens~2 through a ROS~2 middleware framework. Within this ecosystem, ROS~2 is employed to subscribe to the robot's transform (TF) data and command goal poses. The HoloLens~2 serves as the primary interface for capturing and processing user information, which is then utilized to determine and transmit the target pose. To facilitate natural language interaction, an independent server is implemented to host the Speech-To-Text (STT) and the Large Language Model (LLM) services. Bidirectional communication between the ROS~2 and the HoloLens~2 is established via the ROS-TCP-Connector, while the HoloLens~2 interacts with the STT and LLM servers through standard HTTP requests by sending POST requests.

\begin{figure}[t]
    \centering
    \includegraphics[width=0.9\columnwidth]{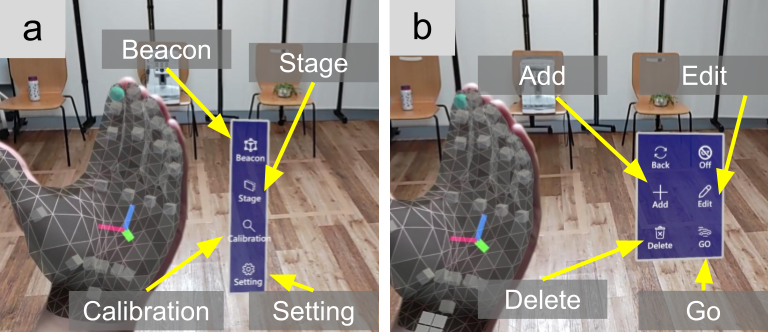}
    \caption{\textbf{Hand Menu.} (a) The primary layout of the hand menu. (b) The Beacon submenu, accessed via the main menu.}
    \label{fig:Hand Menu}
\end{figure}

To access the MR-beacon functionalities for robot navigation, a hand menu is invoked using the "Hand Constraint Palm Up" gesture, as illustrated in Fig.~\ref{fig:Hand Menu}. This menu provides access to three primary menu: the Beacon, Stage, and Calibration buttons. The Stage button is used for experimental control, while the Calibration button allows for the adjustment of the user's minimum audio volume threshold. Upon selecting the Beacon button, the interface expands into six specialized functions: Back, Off, Add, Edit, Go, and Delete. The Back button returns the user to the main menu, and the Off button deactivates all MR-beacon functionalities. The core operations of the system, Add, Edit, Go, and Delete functions are executed through their respective buttons.

\begin{figure}[t]
    \centering
    \includegraphics[width=\columnwidth]{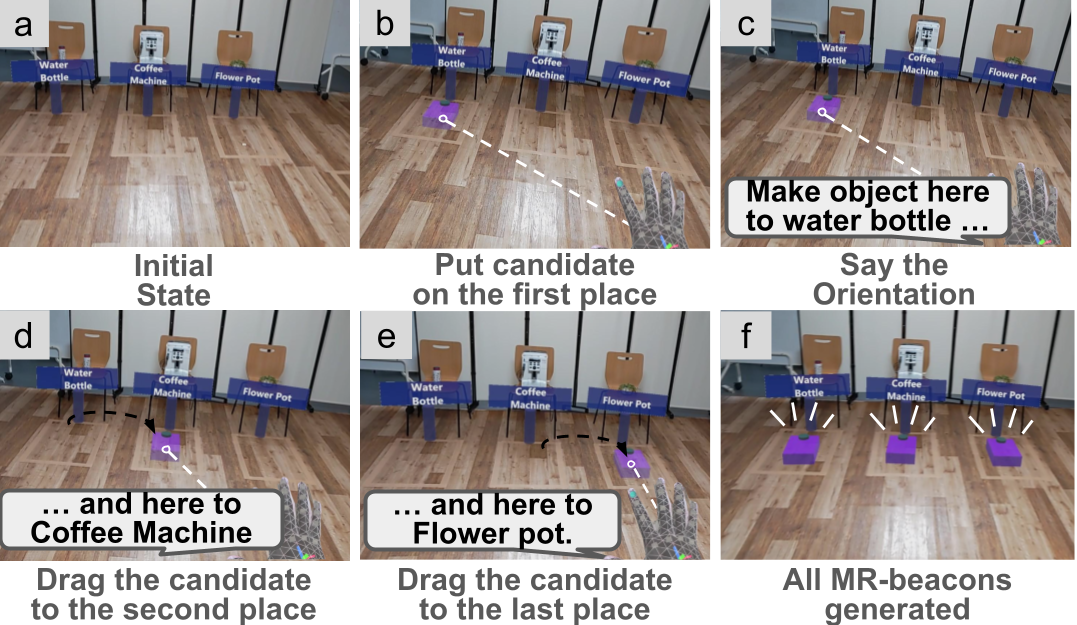}
    \caption{\textbf{Add Function (Multiple Objects).} (a) Initial empty environment. (b) User positioning the first MR-beacon candidate. (c) User specifying orientation toward the water bottle via speech by specifying the name. (d, e) User repeating the same process for the coffee machine and flower pot simultaneously. (f) Final state with three generated MR-beacons. }
    \label{fig:Add2}
\end{figure}

\begin{figure}[t]
    \centering
    \includegraphics[width=\columnwidth]{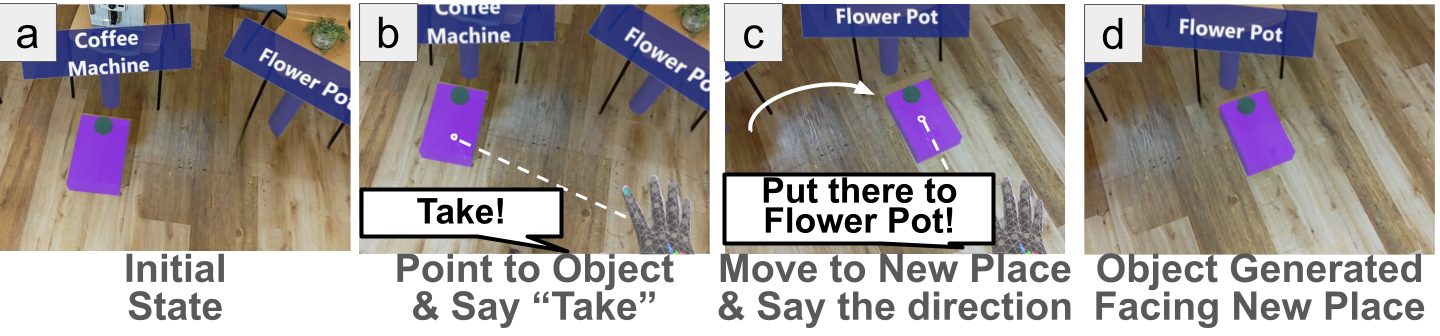}
    \caption{\textbf{Edit Function.} (a) Initial state with an MR-beacon. (b) User selecting the MR-beacon via spatial pointing. (c) User relocating it and specifying a new orientation (flower pot) via speech. (d) Final state of the modified MR-beacon.}
    \label{fig:Edit}
\end{figure}

\begin{figure}[t]
    \centering
    \includegraphics[width=\columnwidth]{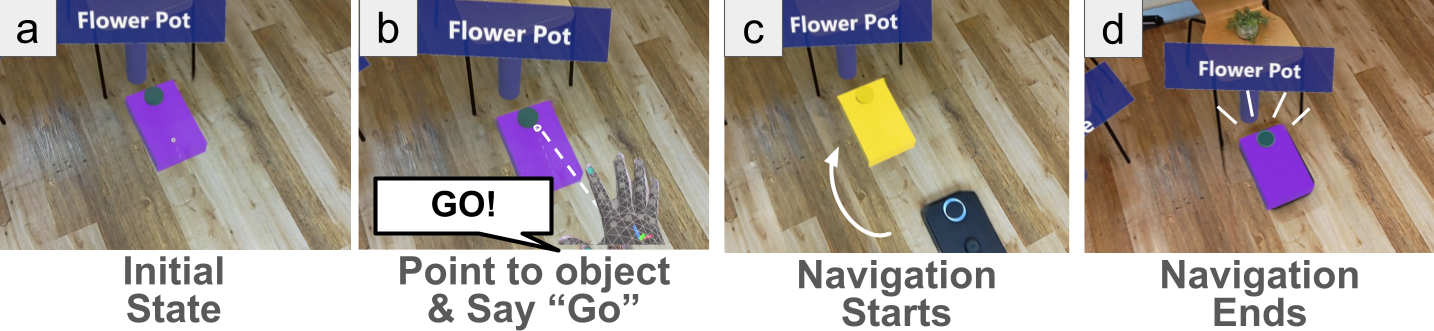}
    \caption{\textbf{Go Function.} (a) Initial state with an MR-beacon. (b) User pointing to the target MR-beacon and issuing the ``Go'' command. (c) Robot navigating to the target. (d) Successful arrival at the destination.}
    \label{fig: GO}
\end{figure}

\begin{figure}[t]
    \centering
    \includegraphics[width=\columnwidth]{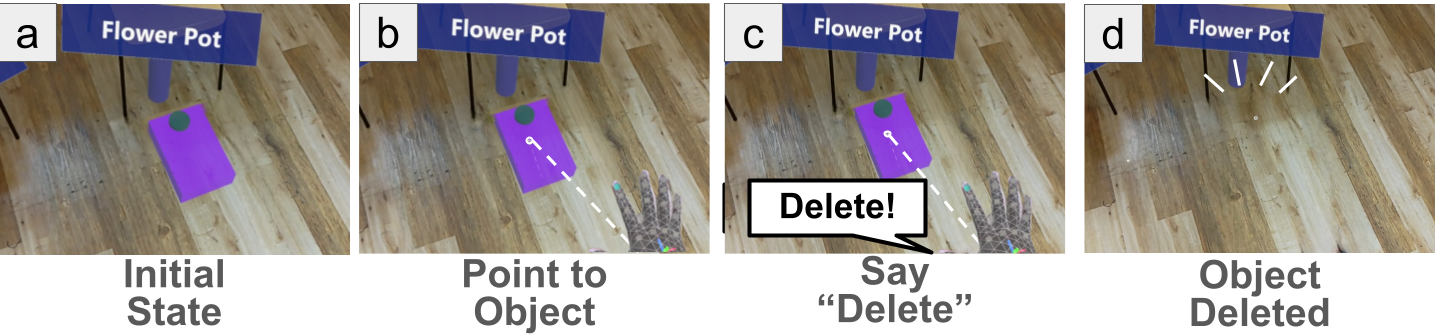}
    \caption{\textbf{Delete Function.} (a) Initial state with an MR-beacon. (b) Pointing to select the target MR-beacon. (c) Issuing the ``Delete'' command to remove the MR-beacon. (d) Final state with the MR-beacon removed.}
    \label{fig: Delete}
\end{figure}
\subsection{Add Function}

The \textit{Add} function generates MR-beacons directly onto the floor plane. Upon pressing the corresponding button, the system enters the ``Add Mode'', which supports the creation of either a single or a sequence of multiple target poses through a continuous pointing and speech workflow.  As shown in Fig.~\ref{fig:Add2}, the user first points to a desired location on the floor, dynamically guiding a semi-transparent candidate MR-beacon in real-time. Upon finalizing the location, the user issues a voice command to specify the robot's orientation, generating a persistent MR-beacon at that pose. For multiple destinations, the user simply maintains this interaction loop, seamlessly shifting the pointer to new locations and issuing corresponding commands for each subsequent target.

\subsection{Edit Function}

The Edit function is designed to modify the spatial coordinates and rotation of an existing MR-beacon. Upon activating this feature, the system transitions into ``Edit Mode'', a two-step procedure comprising beacon selection and repositioning, as shown in Fig.~\ref{fig:Edit}. To select the target, the user directs the spatial pointer toward a previously instantiated MR-beacon. When the cursor intersects the object, the user says ``Take'', effectively attaching the MR-beacon to the pointer's trajectory. To finalize the relocation, mirroring the workflow of ``Add Mode'', the user designates the new target coordinates via spatial pointing while simultaneously saying the robot direction to define the updated orientation.

\subsection{Go Function}

The ``Go'' function navigates the robot to the precise position and orientation of a designated MR-beacon. Pressing the ``Go'' button transitions the system into ``Go Mode'', as illustrated in Fig.~\ref{fig: GO}. First, the user directs the spatial pointer toward an instantiated MR-beacon. Once the cursor intersects the object, the user issues the ``Go'' voice command, prompting the robot to initiate movement toward the target.
\begin{figure}[t]
    \centering
    \includegraphics[width=\columnwidth]{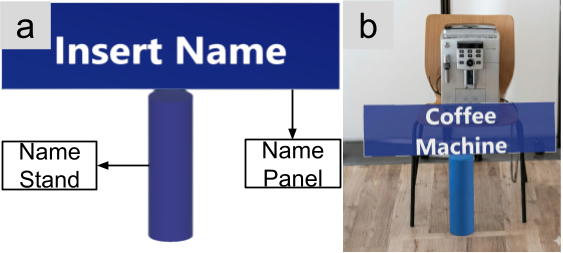}
    \caption{\textbf{MR-label}. (a) illustrates the components of the MR-label, which consists of a name panel to display the location's name and a name stand to mark its position on the floor. (b) demonstrates the deployment of MR-labels in a real-world}
    \label{fig:label}
\end{figure}

\begin{figure}[t]
    \centering
    \includegraphics[width=.8\columnwidth]{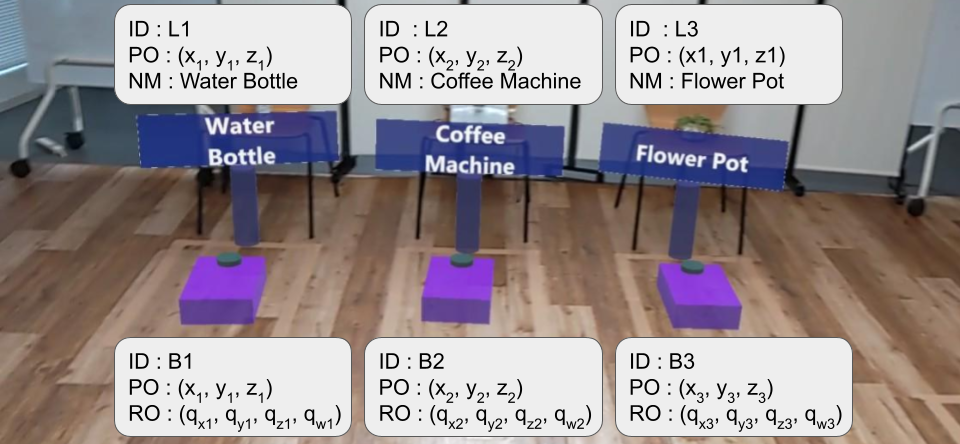}
    \caption{\textbf{Database.} The MR-label database logs the ID, position (PO), and name (NM). The MR-beacons database logs the ID, position (PO), and rotation (RO).}
    \label{fig:DB}
\end{figure}

\subsection{Delete Function}

The ``Delete'' function is used to remove an existing MR-Beacon from the environment. Pressing the ``Delete'' button transitions the system into ``Delete Mode'', as illustrated in Fig.~\ref{fig: Delete}. First, the user aims the spatial pointer at a previously instantiated MR-Beacon. Once the cursor intersects the target object, the user issues the ``Delete'' voice command, which immediately removes the beacon from the physical space.

\section{System Implementation}

\subsection{MR-label}

The MR-label is utilized within the system to assign a unique semantic name to a specific coordinate in the physical environment. This designated location then functions as a persistent directional reference, which the system uses to determine the exact orientation the MR-beacon must adopt upon the generation. The specific structure of these MR-labels and their implementation within the application interface are illustrated in Fig.~\ref{fig:label}.
\label{sec:analysis}
\begin{algorithm}[t]
\caption{User Input Analysis}
\label{alg:user_input_analysis}
\footnotesize
\begin{algorithmic}[1]
\State \textbf{Input:} Hand Menu, Input Acquirer, Label DB
\State $currentMode \gets \text{Hand Menu}$

\If{$currentMode$ is Inference Action}
    \State $(voiceData, pointerData) \gets \text{Input Acquirer}$
    \State $labelNameList \gets \text{AnalyzeVoice}(voiceData)$
    \State $clusterList \gets \text{AnalyzePointer}(pointerData)$
    \State $labelInfo \gets \text{Label DB}$
    \State $\text{CalculatePose}(labelNameList, clusterList, labelInfo)$
\Else
    \State $pointerData \gets \text{Input Acquirer}$
    \State $beacon \gets \text{Identify MR Beacon}(pointerData)$
    \State $\text{Execute Command}(beacon, currentMode)$
\EndIf
\end{algorithmic}
\end{algorithm}

\begin{algorithm}[t]
\caption{Clustering Algorithm}
\label{alg:sequential_clustering}
\footnotesize
\begin{algorithmic}[1]
\State \textbf{Input:} pointList
\State $M \gets \text{GetSize}(pointList)$

\For{$i \gets 1$ \textbf{to} $M$}
    \State $curPoint \gets pointList[i]$
    
    \If{ $Dist(curPoint, curCentroid) <= D_{th}$}
        \State $\text{Recalculate Cluster Centroid}$
        \State $\text{Cluster Size Increases by 1}$
    \Else
        \State $\text{Record Current Cluster}$
        \State $\text{Create New Cluster Size 1}$
    \EndIf
\EndFor

\State $\text{Record Last cluster}$
\end{algorithmic}
\end{algorithm}

\subsection{Database}

Separate databases are implemented for both the MR-labels and the MR-beacons to ensure persistent access to their spatial and semantic data. For the MR-labels, the database stores three primary attributes: (1) ID, a unique identifier generated via a Globally Unique Identifier (GUID), (2) Name, the semantic annotation displayed on the object’s name panel, and (3) Location, a 3D vector $(x, y, z)$ representing the coordinates relative to the global image target.

Similarly, the MR-beacons are stored with the following attributes: (1) ID, a unique identifier generated using a GUID, (2) Location, a 3D vector $(x, y, z)$ defining the MR-beacon's position, and (3) Rotation, a vector or Quaternion $(x, y, z, w)$ representing the MR-beacon's orientation. All spatial data for both databases is calculated and stored relative to the global image target to maintain coordinate consistency.

\subsection{Input Acquirer}
\label{sec:input_acq}

To provide a seamless user experience without requiring manual triggers, we implemented a Voice Activity Detector (VAD). This component monitors audio input to detect the onset of speech (voice detection) and the completion of a command (silence detection) using a predefined threshold.

Once the voice detection is triggered, the system simultaneously captures the audio stream and the user's current spatial pointer coordinates on the floor. To ensure high precision for the pointer analysis process, the pointer data is sampled every 0.1 seconds. After the silence detection is triggered, the system then analyzes the collected voice and pointer data to be further processed based on the current mode.

\subsection{User Input Analyzer}
\label{sec:input_analyzer}

The user input analysis algorithm is described in Alg.~\ref{alg:user_input_analysis}. Based on the current mode from the hand menu, the system categorizes the expected actions from the user into two types, which are the Non-Inference and Inference actions.

Non-Inference actions (e.g., Delete and Go functions) are deterministic and do not require an LLM model, which means they rely solely on the user's pointer coordinate. These actions requires system to detect which MR-beacon did it collide and do the process according to the mode.

Inference actions (e.g., Add Mode) required both synchronized voice and pointer data. This data is processed independently through Voice Analyzer and Pointer Analyzer, before the system integrates these outputs with a predefined MR-label list from the database to calculate the corresponding pose.

\subsubsection{Voice Analyzer}
\label{ch:voice_analyzer}

Voice Analyzer requires two primary components: an STT model and an LLM model. The STT model uses \textit{faster-whisper} with \textit{small.en} model in order to convert raw audio data into text, which is then processed by the LLM to identify the MR-labels for each MR-beacons as a JSON-formatted string. The string is subsequently parsed by a JSON converter to a list of the MR-label name, which is then used to calculate each of the MR-beacons' pose. The model we used for LLM model is \textit{meta-llama/Llama-3.1-8B-Instruct}. 

\begin{algorithm}[t]
\caption{MR-beacon Pose Calculation}
\label{alg:beacon_pose_calculation}
\footnotesize
\begin{algorithmic}[1]
\State \textbf{Input:} labelNameList, clusterList, labelInfo
\State $N \gets \text{GetSize}(labelNameList)$
\State $biggestCluster \gets \text{GetTopNBiggestCluster}(clusterList)$
\State $locationList \gets \text{SortByTime}(biggestCluster)$

\State $AllPose \gets \text{empty list}$

\For{$i \gets 1$ \textbf{to} $N$}
    \State $curLoc \gets LocationList[i]$ 
    \State $curLabel \gets labelNameList[i]$
    \State $curLabelLoc \gets \text{GetCurrentLabelInfo}(curLabel, labelInfo)$
    \State $curRot \gets \text{FindAtan}(curLoc, curLabelLoc)$ 
    \State $curPose \gets \text{ConstructPose}(curLoc, curRot)$ 
    \State Add $curPose$ to $AllPose$
\EndFor

\State \Return $AllPose$
\end{algorithmic}
\end{algorithm}

\subsubsection{Pointer Analyzer}
\label{ch:pointer_analyzer}
Pointer Analyzer analyzes the user's spatial pointer information to determine the location of all possible MR-beacons. The algorithm used to determine the cluster from the pointer data is shown in Alg.~\ref{alg:sequential_clustering}.

The algorithm processes data points one by one, identifying and classifying them by maintaining a running average of the current cluster as its centroid. For each new point, the algorithm checks whether the point is inside the current centroid by calculating its distance from the active centroid whether it is within a certain threshold $D_{th}$. If the point falls within, it is included in the cluster, updating the cluster centroid again while increasing the size of the cluster by 1. Meanwhile, if a point falls outside, the algorithm records the current cluster, saves its final centroid to a list and size, and immediately starts a new cluster using the outlier as the first point. The extracted cluster data is then used to calculate the pose of each of the MR-beacons. This procedure maintains a linear time complexity of $O(n)$.

\subsubsection{Pose Calculation} Finally, the system determines the quantity, position, and rotation of the MR-beacons by integrating the voice analyzer's MR-label list with the pointer analyzer's clusters. As shown in Alg.~\ref{alg:beacon_pose_calculation}, first, the number of beacons to be generated, $N$, is defined by the quantity of detected MR-labels from voice analyzer. To capture the relevant spatial data while preserving the user's intent sequence, the system extracts the $N$ largest pointer clusters and sorts them chronologically. For each $i$-th MR-beacon (from $1$ to $N$), its position is directly assigned to the $i$-th sorted cluster's location. Its rotation (yaw) is oriented toward the corresponding MR-label by calculating the $\operatorname{atan2}$ of the displacement vector between the MR-beacon's position and the MR-label's database coordinates. Finally, the system outputs these combined positions and rotations as target poses to be instantiated as the MR-beacon.

\subsection{Unity-ROS Integration}
\subsubsection{Communication with ROS~2 and Model}
A ROS-TCP-Connector is employed to facilitate communication between the HoloLens 2 and ROS~2. This bridge enables the exchange of robotic data, including Transform (TF) data for spatial co-localization and goal pose data for autonomous navigation.

To maintain a modular architecture, the STT and LLM models are hosted on an independent server using the FastAPI framework via the HTTP protocol, by sending POST requests containing the audio file for the STT model and the transcribed text for the LLM to retrieve the MR-label names.
\subsubsection{Co-Localization}
To align the coordinate systems of ROS~2 and the HoloLens~2, a co-localization process is required. Utilizing Vuforia, a reference object (QR Code) is registered in the Vuforia database to act as the reference point for all MR-objects in HoloLens~2 and also corresponds to the ROS map origin. This results in a simplified coordinate transformation between both coordinate systems.

\begin{figure}[t]
    \centering
    \includegraphics[width=\columnwidth]{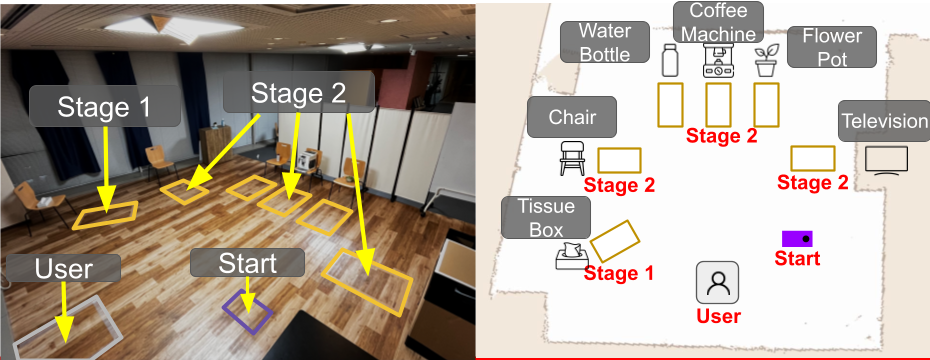}
    \caption{\textbf{Experiment Environment}. (a) shows the figure of the real-world experiment environment. (b) shows the figure of the experiment environment from 2D map SLAM.}
    \label{fig:Experiment_Environment}
\end{figure}
\section{Experiments}
To evaluate the effectiveness of the proposed system, we conducted an experiment comparing the proposed MRPoS system with MRNaB \cite{mrnab} as the gesture-based MR system baseline. The experiment was conducted twice, once for the MRNaB and MRPoS systems per participant, in a counterbalanced order. Sixteen participants were involved in the experiments. Most of whom had no prior experience with ROS or VR/AR/MR.

\begin{figure*}[t]
    \centering
    \includegraphics[width=\textwidth]{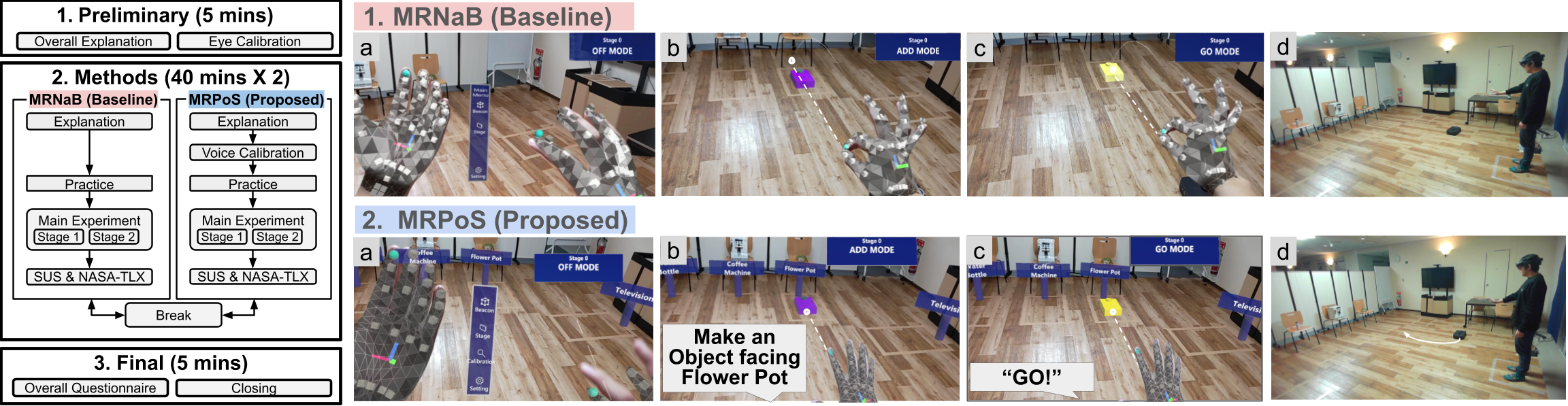}
    \caption{\textbf{Overall Experiment Flow} comparing MRNaB and MRPoS. Upper image shows the experiment flow while the lower image shows the main experiment flow. Both groups access the hand menu to generate the initial objects. Next, MRNaB employs hand gestures to instantiate and select MR-beacon. In contrast, MRPoS utilizes speech and spatial pointing to create and select MR-beacon. Once selected, robot starts the navigation.}
    \label{fig:combined_flow}
\end{figure*}

\subsection{Experiment Setup}
Fig.~\ref{fig:Experiment_Environment} illustrates the experimental environment, which was designed to simulate a typical household setting. The environment includes common items such as a tissue box, chair, water bottle, coffee machine, flower pot, and television. The task was to navigate the robot to predetermined areas, marked by tape with the direction facing to the closest object. 

The experiment consisted of two distinct stages. Stage 1 required participants to navigate to a single target, the tissue box, positioned at an oblique angle of 210°. This stage was specifically designed to evaluate the system’s performance in a challenging orientation that complicates MR-beacon generation. In contrast, Stage 2 involved a sequential navigation task to three locations: the chair, coffee machine, and TV. These targets were positioned at orthogonal angles (0° or $\pm$90°) to represent the more accessible configurations. This stage compared the proposed system’s efficiency against the baseline for multi-destination generation.

As shown in Fig.~\ref{fig:combined_flow}, both the baseline, MRNaB, and the proposed method began with a preparation phase, where participants watched an explanatory video and practiced with the system to familiarize themselves with its operation. Specifically for the proposed method, we also conducted voice calibration using example prompts with the format: ``Make an object here facing \{Place Name\}''. However, they were permitted to customize the phrasing themselves. For the clustering logic in Section~\ref{ch:pointer_analyzer}, the distance threshold $D_{th}$ was set to 0.15m to ensure stable MR-beacon generation. Fig.~\ref{fig:combined_flow} also shows the main experiment of each methods by utilizing each modalities. Finally, each method experiment was concluded with the SUS and NASA-TLX questionnaires. At the end, participants filled in the overall questionnaire.

\subsection{Evaluation Indices}
The evaluation metrics were categorized into two primary groups: objective and subjective indices. The objective metrics comprised three performance indicators: (1) the total time required for participants to generate all MR-beacons, (2) the cumulative count of \textit{Add} and \textit{Edit} actions performed to complete each stage, and (3) the location and rotation errors per MR-beacons from the predefined ground truth.

The subjective indices consisted of two primary metrics. First, system usability with System Usability Scale (SUS) questionnaire \cite{SUS} and second, the system workload, measured by the NASA Task Load Index (NASA-TLX) \cite{NASA}.

To evaluate statistical significance, we first performed the Shapiro-Wilk test \cite{shapirotest} to determine whether our data followed a normal distribution. For datasets that satisfied the normal distribution, we employed a Paired-Sample T-test \cite{pairtest}, while for data that deviated from a normal distribution, we used the Wilcoxon Signed-Rank test \cite{wilcoxon}.

\subsection{Experiment Result}
\begin{figure}[t]
    \centering
    \begin{subfigure}[b]{0.48\columnwidth}
        \centering
        \includegraphics[width=\textwidth]{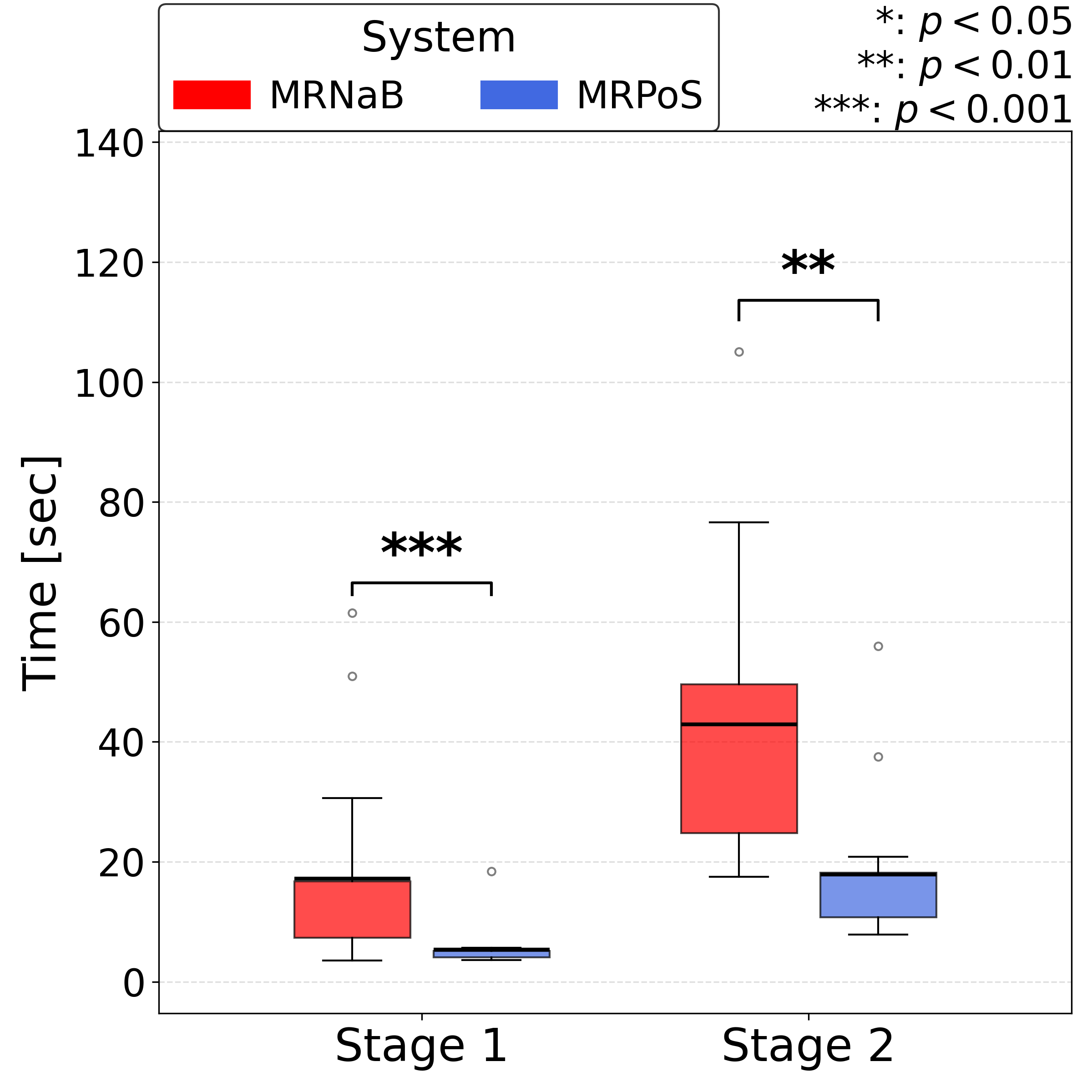}
        \caption{Time Used for MR-beacon}
        \label{fig:time_beacon}
    \end{subfigure}
    \hfill 
    \begin{subfigure}[b]{0.48\columnwidth}
        \centering
        \includegraphics[width=\textwidth]{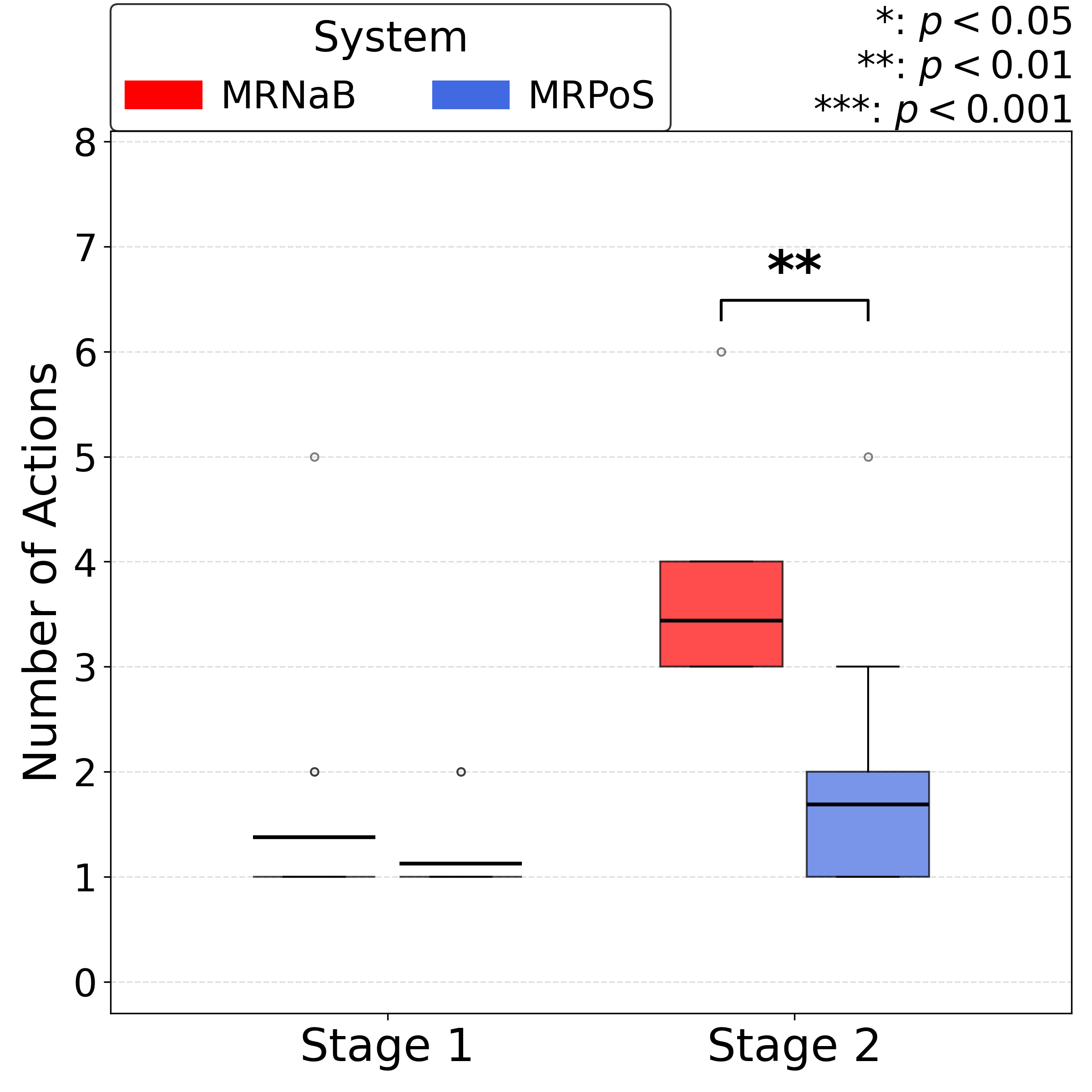}
        \caption{Total Action Number}
        \label{fig:action_count}
    \end{subfigure}
    
    \caption{\textbf{Time and Action Count Result}. Statistical difference on Stage 1 \& 2 for time and on Stage 2 for action.}
    \label{fig:combined_time_action}
\end{figure}
\begin{table}[t]
\centering
\caption{Time and Action Count Results}
\label{tab:combined_results}
  \resizebox{0.85\columnwidth}{!}{
\begin{tabular}{lcccc}
\toprule
& \multicolumn{2}{c}{\textbf{Time [sec]}} & \multicolumn{2}{c}{\textbf{Action Count}} \\
\cmidrule(lr){2-3} \cmidrule(lr){4-5}
\textbf{Stage} & \textbf{MRNaB} & \textbf{MRPoS} & \textbf{MRNaB} & \textbf{MRPoS} \\
\midrule
Stage 1 & 17.21 & \textbf{5.37} & 1.38 & \textbf{1.12} \\
Stage 2 & 42.92 & \textbf{17.88} & 3.44 & \textbf{1.69} \\
\bottomrule
\end{tabular}
}
\end{table}
\begin{figure}[t]
    \centering
    \begin{subfigure}[b]{0.48\columnwidth}
        \centering
        \includegraphics[width=\textwidth]{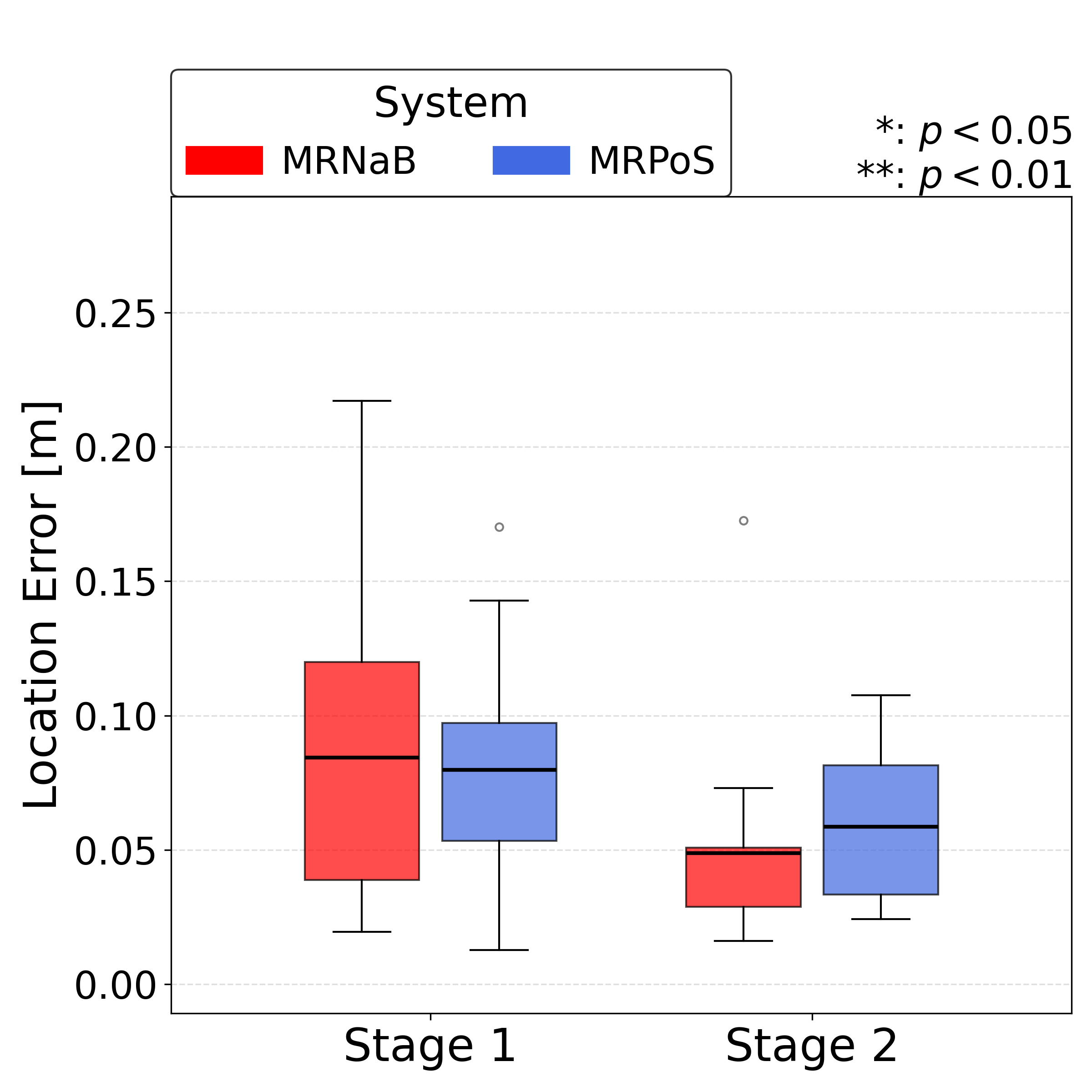}
        \caption{Location Error}
        \label{fig:loc_error}
    \end{subfigure}
    \hfill 
    \begin{subfigure}[b]{0.48\columnwidth}
        \centering
        \includegraphics[width=\textwidth]{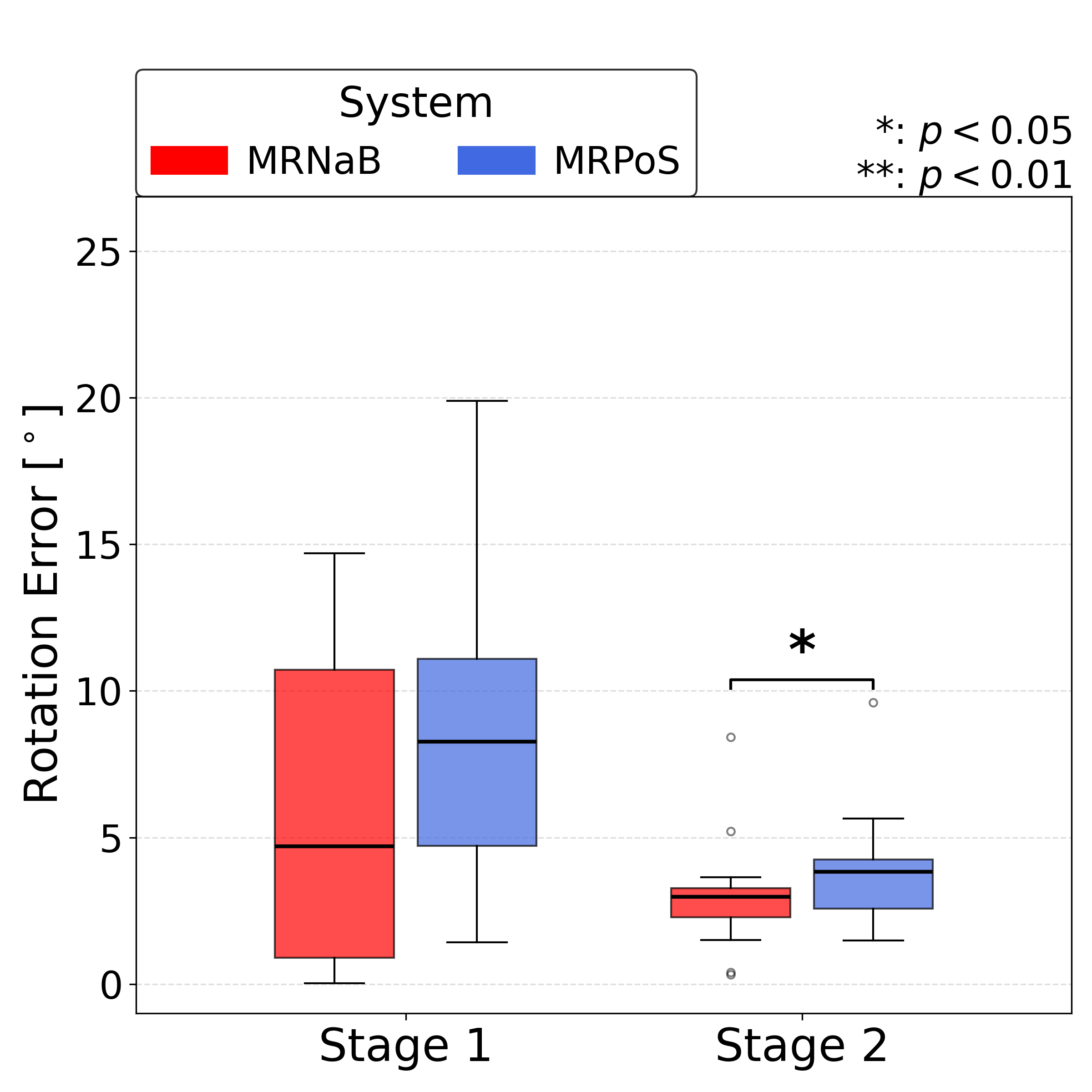}
        \caption{Rotation Error}
        \label{fig:rot_error}
    \end{subfigure}
    
    \caption{\textbf{Location and Rotation Error per MR-beacon Result}.}
    \label{fig:combined_errors}
\end{figure}
\begin{table}[t]
\centering
\caption{Location and Rotation Error}
\label{tab:error_summary}
  \resizebox{0.85\columnwidth}{!}{
\begin{tabular}{lcccc}
\toprule
& \multicolumn{2}{c}{\textbf{Location Error [m]}} & \multicolumn{2}{c}{\textbf{Rotation Error [$^\circ$]}} \\
\cmidrule(lr){2-3} \cmidrule(lr){4-5}
\textbf{Stage} & \textbf{MRNaB} & \textbf{MRPoS} & \textbf{MRNaB} & \textbf{MRPoS} \\
\midrule
Stage 1 & 0.084 & \textbf{0.080} & \textbf{4.70} & 8.27 \\
Stage 2 & \textbf{0.049} & 0.059 & \textbf{2.98} & 3.83 \\
\bottomrule
\end{tabular}
}
\end{table}

\begin{figure}[t]
    \centering
    \begin{subfigure}[b]{0.49\columnwidth}
        \centering
        \includegraphics[width=\textwidth]{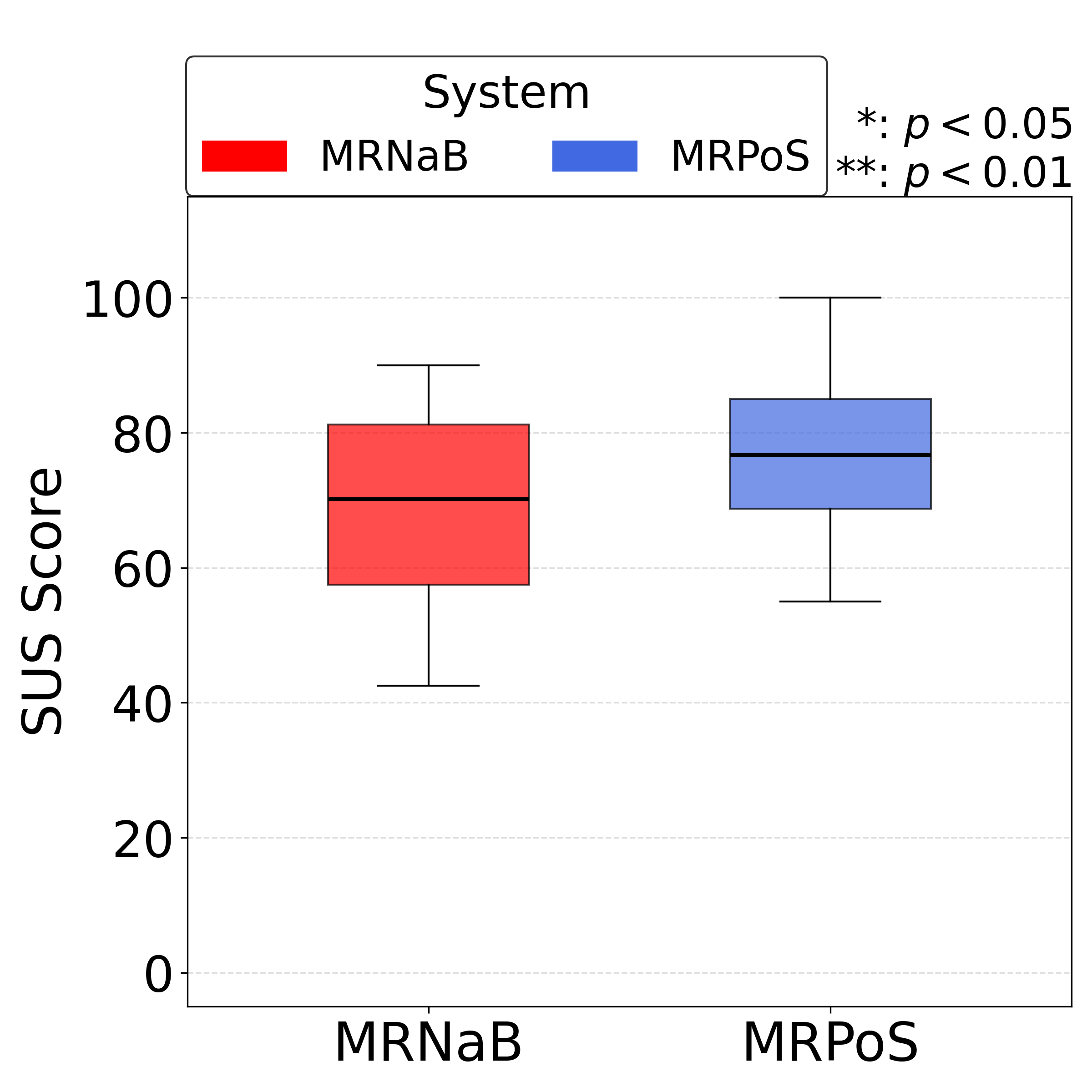}
        \caption{SUS}
        \label{fig:sus}
    \end{subfigure}
    \hfill 
    \begin{subfigure}[b]{0.49\columnwidth}
        \centering
        \includegraphics[width=\textwidth]{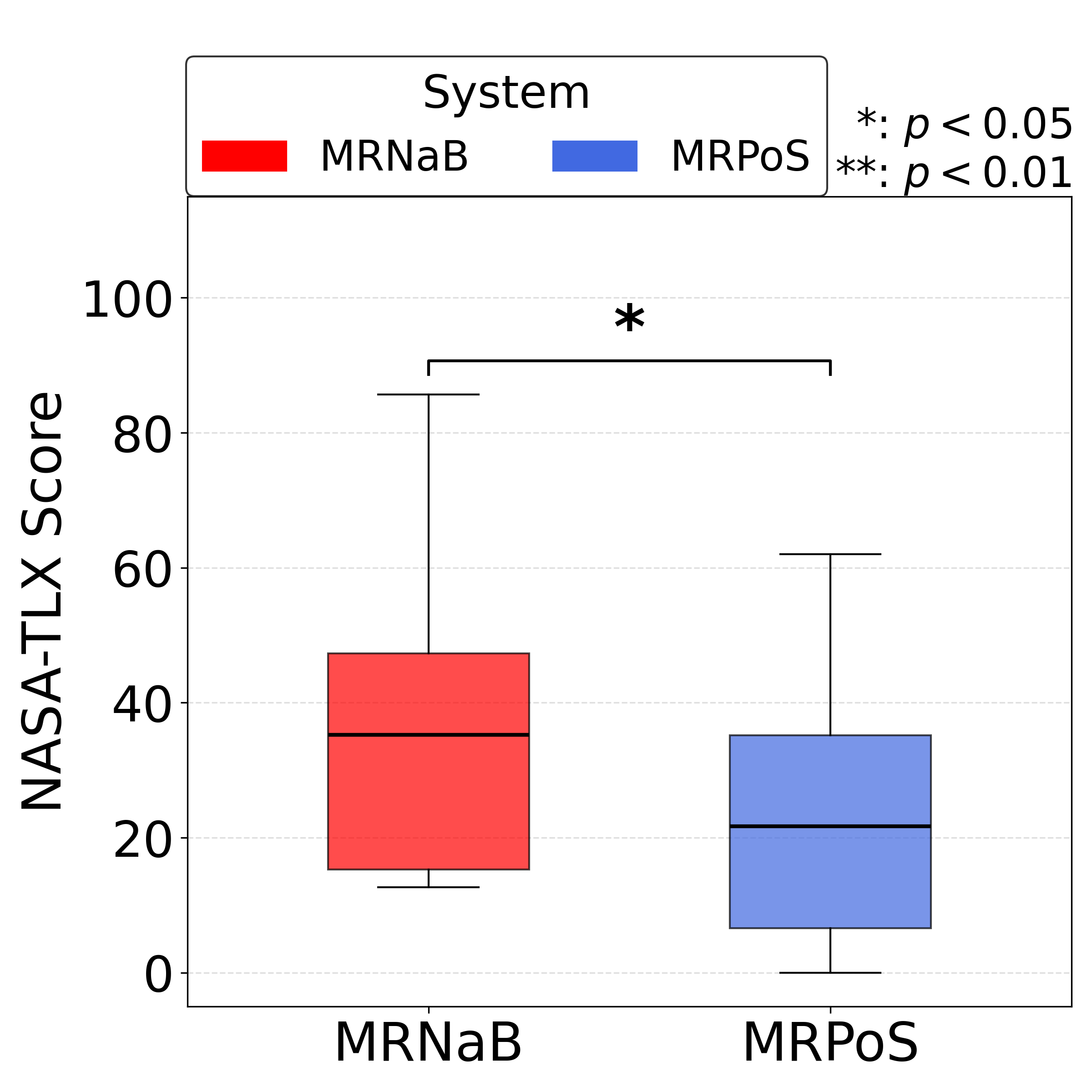}
        \caption{NASA-TLX}
        \label{fig:nasa_error}
    \end{subfigure}
    
    \caption{\textbf{SUS and NASA-TLX Result}. Statistical difference can be found for NASA-TLX}
    \label{fig:subjective}
\end{figure}
\begin{table}[t]
\centering
\caption{Statistical Analysis of Subjective Metrics (SUS and NASA-TLX)}
\label{tab:subjective_metrics}
  \resizebox{\columnwidth}{!}{
\begin{tabular}{lcccc}
\toprule
\textbf{Metric} & \textbf{MRNaB} & \textbf{MRPoS} & \textbf{$p$-value} & \textbf{Effect Size} \\
\midrule
SUS Score $\uparrow$    & 70.16 & \textbf{76.72} & 0.064 & 0.464 \\
NASA-TLX $\downarrow$   & 35.27 & \textbf{21.67} & 0.013 & 0.620 \\
\bottomrule
\end{tabular}
}
\end{table}

\begin{table}[t!]
\centering
\caption{MRPoS Time Breakdown [sec]}
\label{tab:MRPoS}
  \resizebox{\columnwidth}{!}{
\begin{tabular}{lccccc}
\toprule
\textbf{Stage} & \textbf{Voice} & \textbf{STT} & \textbf{LLM} & \textbf{Clustering} & \textbf{Overall} \\ \midrule
Stage 1 & 4.20 & 0.35 & 0.81 & 0.01 & \textbf{5.37} \\
Stage 2 & 15.72 & 0.75 & 1.39 & 0.02 & \textbf{17.88} \\ \bottomrule
\end{tabular}
}
\end{table}

\subsubsection{Time Used to Create Beacon}
The results, shown in Fig.~\ref{fig:time_beacon}, demonstrate a statistically significant reduction in task completion time ($p = 0.00076$ for Stage 1, $p = 0.0012$ for Stage 2). As shown in Table~\ref{tab:combined_results} (left), participants using the proposed method finished the tasks significantly faster than those using the baseline. On average, our system required only 5.36 seconds for Stage 1 and 17.86 seconds for Stage 2. Conversely, the baseline system averaged 17.21 seconds and 42.92 seconds for Stage 1 and 2. We posit that this substantial increase in time stems from our interface enabling both single-step goal pose generation and the concurrent placement of multiple MR-beacons.

\subsubsection{Action Count Needed}
The results, shown in Fig.~\ref{fig:action_count}, demonstrate a reduction in the required number of \textit{Add} and \textit{Edit} as the primary interactive effort, especially in multiple places condition, achieving statistical significance in the second stage ($p = 0.0022$ for Stage 2). As shown in Table~\ref{tab:combined_results} (right), participants using the proposed method required fewer actions to create the MR-beacons than those using the baseline. On average, our system required only 1.12 actions for Stage 1 and 1.69 actions for Stage 2. Conversely, the baseline system averaged 1.38 actions and 3.44 actions for the respective stages. We posit that this substantial decrease in required actions stems from our interface enabling the concurrent placement of multiple MR-beacons.

\subsubsection{Location and Rotation Error per MR-beacon}

The results, shown in Fig.~\ref{fig:combined_errors}, demonstrate a statistically significant difference only for rotation error during Stage 2 ($p = 0.013$). As summarized in Table~\ref{tab:error_summary}, proposed method yielded a significant difference compared to the baseline MRNaB system exclusively in rotation. We attribute these results to the fact that, for MRPoS, rotation is derived from both the MR-beacon and MR-label locations introducing additional variables which increases error in rotation.

\subsubsection{SUS and NASA-TLX}
The subjective evaluation results, illustrated in Fig.~\ref{fig:subjective} and summarized in Table~\ref{tab:subjective_metrics}, demonstrate a statistically significant reduction in user workload for the proposed MRPoS system compared to the MRNaB baseline ($p = 0.013$). This improvement represents a large effect size ($r = 0.620$), suggesting a substantial decrease in the cognitive and physical demands of the task. Furthermore, the SUS scores indicated a marginal trend toward improved usability ($p = 0.064$) with a substantial medium-to-large effect size ($r = 0.464$). We attribute these findings to the multimodal nature of MRPoS by integrating speech and pointing, the system requires less time and fewer discrete actions while eliminating the need for complex hand gestures, thereby lowering the overall barrier to interaction.

\subsection{Discussion}
Table~\ref{tab:MRPoS} shows that the proposed method outperformed the baseline in both completion time and action count. This improvement stems from the LLM's ability to resolve degraded STT output, such as "An object here facing a Tish box", which would otherwise fails. The decrease in time also means less time to raise hands resulting in lower workload shown in Table~\ref{tab:subjective_metrics}. Table~\ref{tab:MRPoS} demonstrates that the combined STT and LLM latency remains within 1–2 seconds, varying slightly with the complexity of the input. This low-latency performance minimizes the perceived delay, facilitating real-time responsiveness for the MRPoS interface.

Another notable result in Table~\ref{tab:error_summary} is the difference in rotational error between MRNaB and MRPoS, which was higher in Stage 1 than in Stage 2. This discrepancy is largely attributable to the experimental setup of Stage 1. Despite only a single target, the location was positioned at an oblique angle (210°) in an area with limited visibility, which likely reduced participants' precision when matching with the MR-label to create MR-beacons. However, even with this slight rotational variance, the proposed method proved to be significantly more efficient overall.
\section{Conclusion}
This paper proposes MRPoS, a Mixed Reality-based robot navigation interface that leverages spatial pointing and LLM-processed speech to simultaneously define the positions and orientations of multiple MR-beacons. The system features four primary functions, which are Add, Edit, Go, and Delete, enabling users to seamlessly generate single or multiple MR-beacons in a single action, modify their location and rotation, transmit the MR-beacon data to the physical robot to initiate autonomous navigation, and remove unwanted MR-beacons.

To evaluate MRPoS, we conducted a comparative experiment against the ``air tap'' based MR interface, MRNaB. Objective results showed that MRPoS achieved faster task completion with significantly fewer attempts. Furthermore, subjective evaluations confirmed a lower mental workload, validating MRPoS as a more efficient navigation interface.


\vfill

\end{document}